\documentclass[conference]{IEEEtran}
\IEEEoverridecommandlockouts

\usepackage{graphicx}   
\usepackage{cite}       
\usepackage{amsmath,amssymb}
\usepackage{hyperref}
\usepackage{array}
\usepackage{caption}
\usepackage{algpseudocode}

\usepackage{titlesec}
\titlespacing*{\subsection}{0pt}{1ex}{1ex}

\usepackage[dvipsnames]{xcolor}
\usepackage{graphicx}
\usepackage{multirow}
\newcommand{\fon}[1]{\fontfamily{#1}\selectfont}
\usepackage[most]{tcolorbox}
\usepackage{caption}
\usepackage{newfloat}

\begin{document}

\title{Semantic Intelligence: Integrating GPT-4 with A* \\
Planning in Low-Cost Robotics}

\author{Jesse Barkley$^{1}$, Abraham George$^{1}$, and Amir Barati Farimani$^{1}$%
\thanks{\makeatletter\footnotesize%
$^{1}$With the Department of Mechanical Engineering, Carnegie Mellon University,%
{\ttfamily\{jabarkle, aigeorge, afariman\}@andrew.cmu.edu}%
\makeatother}
}
\maketitle

\begin{abstract}
Classical robot navigation often relies on hardcoded state machines and purely geometric path planners, limiting a robot’s ability to interpret high-level semantic instructions. In this paper, we first assess GPT-4’s (Generative Pre-Trained Transformer 4) ability to act as a path planner compared to the A* algorithm, then we present a hybrid planning framework that integrates GPT-4’s semantic reasoning with the A* algorithm on a low-cost robot platform operating on ROS2 Humble. Our approach eliminates explicit finite state machine (FSM) coding by using prompt-based GPT-4 reasoning to handle task logic while maintaining the highly accurate paths computed by A*. The GPT-4 module provides semantic understanding of instructions and environment descriptors (e.g., recognizing toxic obstacles or crowded areas to avoid, or understanding low-battery situations requiring proper route selection), and dynamically adjusts the robot’s occupancy grid (via obstacle “buffering”) to enforce semantic constraints. We demonstrate multi-step reasoning for sequential tasks, such as first moving to a resource goal and then navigating to a final goal safely. Experiments on a Petoi Bittle robot (with overhead camera streaming camera feed and Raspberry Pi Zero 2W on the robot) compare classical A* against GPT-4-assisted planning. Results show that while A* is faster and more accurate for basic route generation and obstacle avoidance, the GPT-4 integrated system achieves high success rates (96–100\%) on semantic tasks that are infeasible for pure geometric planners. This work highlights how affordable robots can exhibit intelligent, context-aware behaviors by leveraging advanced LLM reasoning with minimal additional hardware and no LLM fine-tuning.

\end{abstract}


\section{Introduction}
Autonomous navigation in robotics traditionally separates high-level decision making from low-level path planning \cite{1}. High-level behaviors are often encoded in finite state machines (FSMs) or behavior trees  and low-level collision avoidance is handled by algorithms such as A* \cite{1}. However, hardcoding states for every possible scenario (e.g., “avoid hazardous spill” or “find charger when battery is low”) is labor-intensive and lacks flexibility. Recent advances in large language models (LLMs) such as GPT-4 \cite{2} suggest a new paradigm: using an LLM’s reasoning ability to interpret goals and environmental context, then dynamically selecting from or enhancing classical planner’s generated routes \cite{3, 4, 5}. This integration can endow even low-cost robots with semantic intelligence previously seen only in more advanced systems. Frameworks such as SayCan, LM‑Nav, and ChatGPT‑for‑Robotics \cite{6, 7, 8} demonstrate that these models can leverage world knowledge and contextual reasoning, enabling robots to handle abstract instructions. For example, an LLM can infer that a “crowded area” should be avoided for safety even if it is not explicitly marked on a map. The challenge is how to integrate such semantic reasoning with reliable motion planning without extensive additional computation on the robot. In this work, we demonstrate a hybrid planning pipeline (Figure 1) that combines GPT-4’s \cite{2} reasoning with the A* path planning algorithm \cite{1} on a resource-constrained robot. The robot used is the Petoi Bittle \cite{9}, an affordable quadruped, paired with an overhead camera for localization. The system does not use any hardcoded state machine; instead, GPT-4 creates a planning loop that decides the robot’s task execution through understanding the semantics of carefully designed prompts. GPT-4 interprets high-level instructions (including safety or context cues) and selects or modifies A* paths accordingly by adjusting a “buffer” (expanding obstacles to create larger clearance) in the occupancy grid. It can also execute sequentially, handling multi-step tasks through consecutive reasoning stages \cite{10, 11}.

\subsection{Contributions}
This paper’s key contributions include:
\begin{itemize}
\item \textbf{LLM-A* Integration:} A novel hybrid navigation system where GPT-4’s semantic reasoning is integrated with classical A* planning. The LLM analyzes candidate paths from A* and guides the selection based on high-level criteria, demonstrating the best of both approaches \cite{1, 12}.
\item \textbf{Elimination of Hardcoded FSMs:} Task logic is controlled by GPT-4 prompts rather than fixed state machines. This significantly reduces manual coding of behavior logic; the robot’s behavior adapts based on the prompt and scenario rather than static code \cite{4, 5}.
\item \textbf{Semantic Understanding in Navigation:} We show GPT-4 interpreting semantic cues in the environment given information from an occupancy grid and a fine-tuned YOLOv8 model \cite{13} for object detection. For instance, it recognizes descriptors like “toxic spill” or “crowded region” from textual prompts and ensures the chosen path avoids those areas. It also understands contextual goals such as seeking a battery charging station when the robot is described as low on battery \cite{14}.
\item \textbf{Dynamic Occupancy Grid Buffering:} We introduce dynamic buffer adjustment via GPT-4. The LLM decides when to inflate obstacles in the occupancy grid (e.g., adding a safety margin in congested areas), balancing safety and efficiency \cite{15}. This allows on-the-fly route modulation that pure A* cannot achieve on its own.
\item \textbf{Sequential Task Reasoning:} The system can carry out multi-stage tasks (e.g., first navigate to collect a resource, then proceed to a final destination), building on techniques from TidyBot \cite{16}. GPT-4 manages the sequence through iterative prompting, reasoning about intermediate goals before the final objective \cite{17}.
\item \textbf{Low-Cost Implementation:} All experiments are conducted on a low-cost setup: a Petoi Bittle robot \cite{9} with an overhead web camera and a Raspberry Pi Zero 2W on the robot. We demonstrate that advanced LLM-based intelligence with proper prompting \cite{18} can be brought to hobbyist-level robots, making our approach accessible and reproducible.
\end{itemize}

\begin{figure}[!ht]
\centering
\includegraphics[width=\linewidth]{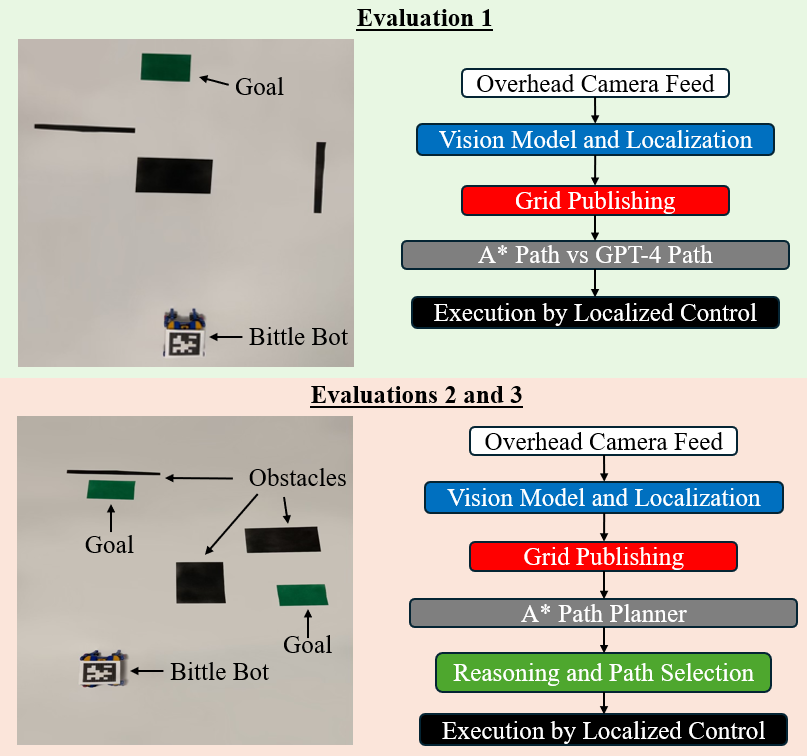}
\caption{Diagram of our evaluations and system design. 
Evaluation 1 tested GPT-4's performance in generating a route 
around obstacles to a desired goal against an A* path planning 
algorithm. Both systems were given identical routes for consistency 
to measure accuracy and time. Three courses were made for this 
assessment, increasing in difficulty from one obstacle to a max of 
three. Evaluations 2 and 3 enabled an A* path planner to generate routes 
to all possible goals, then giving GPT-4 a task to select the 
appropriate route that achieved the desired result of the prompt.}
\label{fig:fig1}
\end{figure}

By leveraging GPT-4’s cognitive abilities in tandem with a reliable path planner, our system achieves robust, context-aware navigation. We evaluate its performance against classical planning and in various semantic scenarios. The results show that the integrated approach retains the strengths of A* for obstacle avoidance while dramatically expanding the range of tasks the robot can handle, all without additional expensive sensors, computing hardware, or state machine code refinement.

\subsection{Related Work}
\textbf{LLMs in Robotic Planning:} Large language models have recently been applied to robotics planning problems for their general reasoning ability. Ahn et al.’s SayCan is a seminal approach that grounds an LLM’s suggestions in the robot’s physical abilities \cite{6, 19}. In SayCan, a language model proposes high-level actions and a value function (learned from robot skills) evaluates which actions are feasible, allowing long-horizon tasks to be executed by combining semantic knowledge with embodied constraints. Zhan et al. introduced Mc-gpt \cite{20}, which combined pre-trained models for language, vision, and navigation to enable instruction-driven navigation without additional training data. LM-Nav uses a GPT-3 based module to interpret natural language commands into waypoints, a vision-language model (CLIP) to align images with landmarks, and a navigation policy (ViNG) for low-level control, demonstrating zero-shot navigation in complex outdoor environments \cite{7}. LLMs used for robotic planning can sometimes be used directly out of the box, though many frameworks incorporate a learned value function, or other pre-trained models. This variety of approaches highlights that semantic reasoning can complement existing motion planners, particularly on robots with limited sensor hardware \cite{21}. 

\textbf{LLM and Classical Planner Integration:} Other works explicitly integrate LLMs with classical planning algorithms. LLM+P \cite{3} incorporates an optimal planner into the LLM loop, while Code‑as‑Policies generates executable low‑level code from language to control embodied agents \cite{22}. These methods convert a problem described in natural language to a PDDL (Planning Domain Definition Language) specification, uses a classical planner (e.g., Fast Downward) to solve for an optimal plan, then translates that plan back into natural language steps. SayPlan grounds LLM reasoning in 3‑D scene graphs for scalable task planning \cite{23}. Safety‑aware task‑planning frameworks show that LLM decisions can be vetted or refined to ensure safe execution \cite{24}. These techniques exploit LLMs for understanding the task while ensuring correctness via classical solvers, yielding optimal solutions on benchmark planning problems that raw LLMs struggled to solve. Our work shares a similar spirit of marrying LLMs with classical algorithms, but focuses on real-time navigation and dynamic environment interpretation, rather than discrete symbolic planning.

\textbf{Prompting Strategies for Robotic Tasks:} Several frameworks have been proposed to better prompt LLMs for generating valid robot action plans. ProgPrompt \cite{25} introduced a programmatic prompt structure, essentially giving the LLM a templated “program” with placeholders for actions and sensor inputs. By listing available actions (functions) and objects in a Python-like syntax, ProgPrompt confines the LLM’s output to feasible actions for a given environment, achieving state-of-the-art success in simulated household tasks. Similarly, Robotic Programmer conditions an LLM on instructional video to synthesize manipulation code \cite{26}. Code‑as‑Policies again exemplifies programmatic prompting for embodied control \cite{22}. These efforts show that with structured prompts and tool use, LLMs can effectively orchestrate low-level skills, an idea we utilize by giving GPT-4 a structured view of candidate paths and their metrics.

\textbf{Safety and Reliability in LLM Planning:} Safety is a prominent concern when deploying LLM-driven decision making in robotics. While not the primary focus of our work, we note approaches like SAFER \cite{27} that introduce a safety supervisor alongside the main LLM planner. In SAFER, multiple LLMs collaborate such that one generates a plan and another (the “Safety Agent”) checks or critiques it, possibly injecting corrections if unsafe actions are detected. An LLM-as-a-Judge mechanism is used to evaluate potential rule or safety violations in the plan, and classical safety controllers (e.g., Control Barrier Functions) are integrated to provide guaranteed safety at execution time. Our system inherently limits unsafe actions by using A* (which will not produce physically infeasible paths) and by encoding basic safety rules in the prompt (e.g., instructing GPT-4 to avoid certain regions or add a buffer if the route is congested). However, as LLMs can occasionally produce inconsistent or incorrect decisions, ensuring safety remains an open topic and techniques like multi-agent verification could further enhance our framework.

In summary, prior work has demonstrated the promise of LLMs for translating natural language into robotic actions, whether through grounding in affordances, integration with planners, or clever prompting. We build on this foundation by applying an LLM (GPT-4) to a real-time navigation task on a physical low-cost robot, specifically focusing on semantic reasoning in path planning. To our knowledge, this is the first demonstration of GPT-4 guiding a classical planner for a mobile robot to handle semantic navigation challenges without any hardcoded logic.

\section{Methodology}
Our system architecture combines a vision-based localization setup with a fine-tuned YOLOv8 model, a classical path planner, and an LLM reasoning module (GPT-4). The overall flow is shown in Figure 2 consists of: an overhead camera provides the global position of the robot and identifies relevant world elements such as goals and obstacles; a 2D occupancy grid map is generated based on those objects; in one evaluation A* is compared to GPT-4 for raw path planning performance; in the next evaluation the A* algorithm generates path candidates on this grid and GPT-4 is given a description of these candidates and high-level instructions, selecting the best path (and adjusts parameters like obstacle buffer if needed); finally, the robot executes the chosen path and is tracked using an AprilTag. We detail each component in Figure \ref{fig:fig2} below.

\begin{figure}[!ht]
\centering
\includegraphics[width=\linewidth]{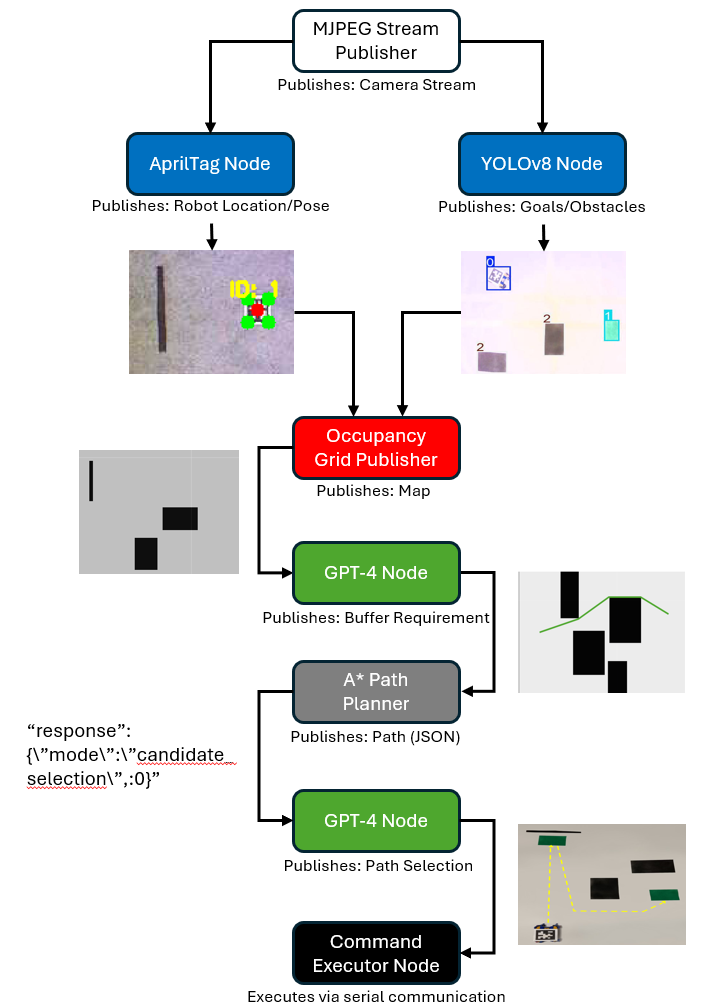}
\caption{Flowchart of the system design and ROS2 nodes 
demonstrating the flow of information to the GPT-4 Node for 
semantic evaluation and route selection given a prompted task.}
\label{fig:fig2}
\end{figure}

\subsection{Hardware Platform}
We use the Petoi Bittle robot, a palm-sized quadruped with servo-driven legs, as our mobile platform. An overhead camera is mounted looking at the robot’s area, tracking the robot’s pose and detecting obstacles and goal markers. The choice of an overhead camera (as opposed to onboard sensors) simplifies the state estimation and map generation, given the limited payload of the Bittle. All computation is handled on a computer and communicated to the Raspberry Pi on the robot through a shared ROS ID. This computer receives camera feed, updates the occupancy grid, runs the A* planner, and communicates with the GPT-4 API for reasoning. Robot locomotion commands (a sequence of waypoints or directional instructions) are sent to Bittle’s microcontroller for execution.

\subsection{Occupancy Grid and A* Planner}
The environment is represented as a 2D occupancy grid with a fixed cell resolution (we used 1 cm per cell in an approximately 1m × 1m arena). Obstacles detected by the overhead camera (black shapes of varying size on the floor) are marked as occupied cells. Free space is marked as free cells. The robot’s starting position and the goal location(s) are identified by an AprilTag on the robot and a green square for the goal(s). The A* algorithm runs on this grid to find collision-free paths from the robot’s current position to a goal cell. 

We extend the basic A* planner to generate candidate paths under different conditions. In classical use, A* would simply output one shortest path to the goal given the current grid. In our system, we sometimes consider multiple goals or multiple clearance settings:
\begin{itemize}
\item \emph{Multiple goals:} If multiple goals are present (for example, a “resource” location and a final destination), A* computes a path to each goal.
\item \emph{Buffer variations:} We also vary the occupancy grid buffer: this is a margin of inflation around obstacles. A buffer of 0 uses the raw grid, while a buffer of $n$ expands each obstacle by $n$ cells (marking those cells occupied as well) to force the planner to give obstacles a wider berth. A larger buffer yields safer but potentially longer paths. We pre-compute paths with different buffer values (e.g., 0 and 20 cm) as separate candidates.
\end{itemize}
Each path candidate is analyzed to extract simple metrics that are useful for GPT-4. We include the path length (distance in cm), the number of turns, the count of obstacles nearby or along the path, and qualitative descriptors of the obstacles arrangement (e.g., “a single narrow obstacle near the goal” or “multiple clustered obstacles forming a narrow passage”). These metrics are packaged into a prompt-friendly format.

\subsection{GPT-4 Prompt Design}
GPT-4 serves as the high-level decision maker. We craft system prompts that describe the robot’s mission and present the candidate path options with their metrics. The prompt is structured in two stages for sequential tasks:

\begin{itemize}
\item \textbf{Stage 1 (if applicable):} If an intermediate target like a resource needs to be collected, GPT-4 is instructed that there are two goals in the environment and it should choose the one corresponding to the resource first. We include distinguishing information (e.g., “the goal with a single skinny obstacle near it is the resource”) so GPT-4 can identify the correct candidate path among those provided. GPT-4 is asked to output a JSON with its choice (\texttt{"selected\_candidate": <index>}) and a buffer setting (\texttt{"buffer": <value>}). The buffer corresponds to whether the path should be taken as-is or recomputed with a safety margin if there are obstacles between the robot and that goal.
\item \textbf{Stage 2:} Once the first stage is complete (or if no intermediate goal, we directly proceed here), GPT-4 is given the task of reaching the final goal safely. The prompt provides any contextual cues (e.g., “there is a toxic spill in the area” or “the route might be crowded”) and lists the candidate path(s) to the final destination. GPT-4 is instructed similarly to choose the best path and decide on a buffer for safety if needed.
\end{itemize}

An example system prompt for the sequential resource collection task is as follows (paraphrased for brevity):

\emph{``You are an LLM controlling a Bittle robot on a 2D grid. The mission is to collect a resource. There are two goals in the environment. The goal with a single skinny obstacle nearby is the resource; that is your target. The other goal should be avoided for now. Based on the candidate path metrics provided (path length, obstacle count, clearances, and qualitative obstacle descriptions), select the candidate that goes to the resource marker. Then decide on the appropriate occupancy grid buffer: if the path is direct with minimal obstacles, set buffer = 0; if it requires turns or obstacle avoidance, set buffer = 20. Respond with JSON: \{"mode":"candidate\_selection", "selected\_candidate":\textless{}index\textgreater, "buffer":\textless{}value\textgreater\}.''}

A similar prompt structure is used for the final navigation stage, emphasizing safety: e.g., “Now navigate to the final goal while maintaining safe clearance. Among the candidates, choose the path that is safest. If the path is sufficiently direct (few obstacles), buffer = 0; if it has narrow passages or obstacles to avoid, buffer = 20.” The prompt approach draws inspiration from instructing GPT-4 as a reasoning engine that weighs options and explicitly outputs a structured decision. We found that framing the task in second person (“You are an LLM controlling a robot…”) and providing clear criteria for decisions helped GPT-4 remain focused and consistent.

\subsection{Execution Loop}
Once GPT-4 returns a decision, the system parses the JSON output. The selected candidate index corresponds to a specific goal (or a specific buffer setting). If GPT-4 suggests a buffer change, the occupancy grid is inflated accordingly and A* is re-run to obtain the final path under those conditions. The robot then receives the waypoints of the chosen path to execute. In sequential tasks, after completing the first leg (resource collection), the system triggers the second stage prompt for the final leg.

It’s important to note that GPT-4 is not generating low-level movement commands; it is only influencing which path the robot takes and how cautiously (via buffer) it should approach it. This keeps the robot’s motion physically safe and feasible, as the actual paths are always generated by the proven A* algorithm on an accurate occupancy grid.

\subsection{Semantic Inputs}
To enable GPT-4’s semantic reasoning, we incorporate high-level descriptors into the prompt. The prompt will state that an area contains a long skinny “TOXIC” obstacle or that an area is “CROWDED” and the LLM must reason as to which goal to select and if a buffer is required based on the data it is receiving from the occupancy grid, path planner, and YOLO. This mechanism is what allows the planner to handle instructions like “avoid the crowded area” that pure geometry-based planners cannot interpret. GPT-4 translates such an instruction into a modification of the path choice (or buffer increase) implicitly.

In summary, our methodology combines:
\begin{itemize}
\item Perception: overhead camera to occupancy grid (ensuring low-cost robust localization and mapping).
\item Classical Planning: A* generates viable path options.
\item LLM Reasoning: GPT-4 ingests path options plus semantic annotations and instructions, and outputs a high-level decision (choice of path, adjustment of safety margin).
\item Control Execution: the robot follows the decided path, completing tasks possibly in stages.
\end{itemize}
This design leverages the strength of both paradigms: the optimality and reliability of classical planners with the flexibility of language understanding. Next, we describe how we evaluated this system through a series of experiments and present the results.

\section{Experiments}
We designed three sets of experiments shown in Figure 3 to evaluate our GPT-4 + A* integrated planner in comparison to a baseline and under various task scenarios. All experiments were conducted in an approximately 1m by 1m indoor arena observed by the overhead camera. We ran multiple trials for each condition to gather quantitative metrics.

\subsection{A* vs GPT-4 Full Control (Geometric Navigation Tests)}
The first experiment compares the performance of a standard A*-based navigation (with no GPT reasoning) to our GPT-4-guided navigation on purely geometric tasks essentially creating a baseline to examine path generation of a traditional planner vs GPT-4. We created three obstacle courses of increasing complexity:

\begin{itemize}
\item \textbf{Course 1:} A single obstacle placed between the start and goal.
\item \textbf{Course 2:} Two obstacles positioned such that the robot must navigate a slight maze to reach the goal.
\item \textbf{Course 3:} Three obstacles creating a more complex winding path to the goal.
\end{itemize}

\begin{figure}[!ht]
\centering
\includegraphics[width=\linewidth]{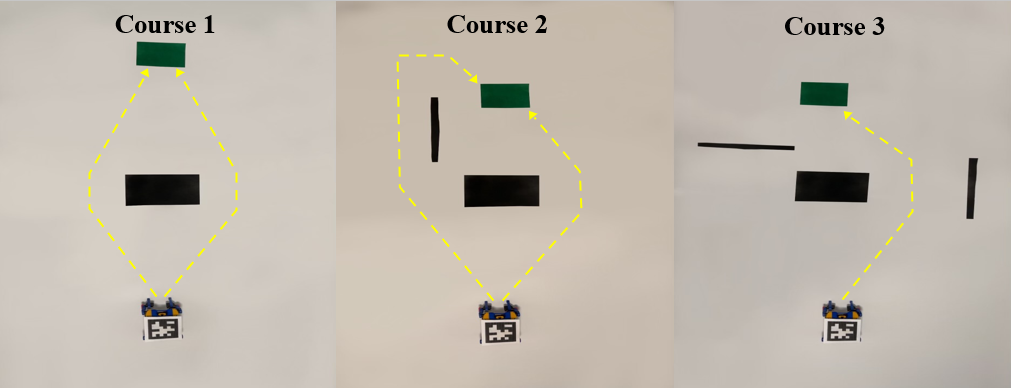}
\caption{Three courses were created to measure the effectiveness of GPT-4 for path planning compared to A*. The courses increased in complexity 
from one obstacle to three, measuring GPT-4's ability to understand the location of multiple obstacles while generating a safe route to the goal.}
\label{fig:fig3}
\end{figure}

In all cases, there were no additional semantic instructions- just “reach the goal” for both approaches. For the GPT-4 trials, we still ran the system through its prompt, but the prompt only had basic geometric info (with no special instructions to avoid any particular object beyond general obstacle avoidance). We measured navigation time (from start to reaching goal), which includes any path planning or reasoning delays, and success rate. A trial is deemed successful if the robot reached the goal without colliding with an obstacle. We also logged the GPT-4 reasoning latency (time taken to receive a response) as part of planning time. To provide GPT-4 more of a chance of success, once waypoints were generated for the LLM’s route a quick check was computed to determine if there would likely be a collision, at which time another API call would occur to generate a new route.

\subsection{Semantic Interpretation Tests (LLM-augmented tasks)}
The second experiment evaluates scenarios where pure geometric planning is insufficient. We crafted three scenarios where an instruction or environmental condition requires semantic understanding:

\begin{itemize}
\item \textbf{Avoid Toxic Obstacle:} In this scenario, one of the obstacles was labeled “TOXIC” and was described to the LLM as a “long skinny obstacle”. The instruction to the robot was to navigate to the correct goal while avoiding the goal near the toxic obstacle. Notably, the shortest path to the goal might involve going near the toxic obstacle. We want to see if GPT-4 will choose a safer path (even if longer) that stays far from the toxic object. We ran 10 trials with varying placements of a “toxic” obstacle and an alternative route.
\item \textbf{Battery Near Goal (Low Battery Scenario):} Here the environment contained two possible goal locations. One goal had the same long skinny obstacle near it that was previously described as “toxic”, but in this case was described as a “charging station”, and the LLM was informed in the prompt that it was “low on battery”. The high-level instruction was to get to the final destination to charge. This tests if GPT-4 will select the goal with the “battery” even though the information the LLM receives from the occupancy grid would indicate it is an obstacle. The LLM had to reinterpret this object as a battery based on the information in the prompt. We ran 10 trials with different arrangements of the “battery” goal versus a regular goal.
\item \textbf{Avoid Crowded Area:} In this scenario with two possible goals, one goal was in a “crowded area” (we placed multiple obstacles around one goal where it was still reachable though more difficult, and the prompt explicitly told GPT-4 that one goal is crowded). The robot should avoid travelling to that goal, even if A* is able to generate a route to the crowded goal. We conducted 10 trials where GPT-4 was to select the uncrowded goal which was deemed safer, testing the LLMs ability to understand goal locations and proximity to obstacles. 
\end{itemize}

For these semantic tasks, our baseline for comparison is not classical A* (which has no mechanism to follow such instructions by itself) but rather the expected correct behavior. We measure semantic success rate, i.e., the percentage of trials in which the robot correctly followed the instruction (avoided the specified region or object, or went to the charging station). We also confirm that the robot still reaches the final goal eventually (task completion).  

\begin{figure*}[!ht]
\centering
\includegraphics[width=\linewidth]{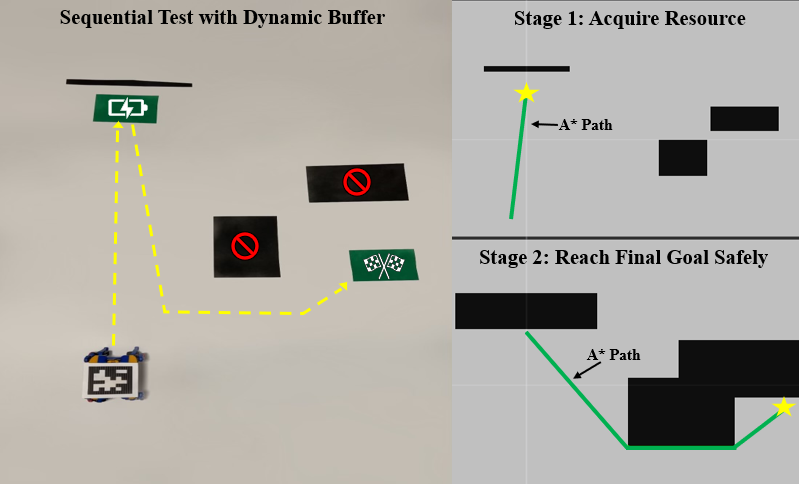}
\caption{Figure 4: LLM executes sequential tasks where the first task is to go to the goal with a "long skinny object that is a resource" (challenging GPT-4 to reinterpret an obstacle as a resource) and the second task is to travel safely to the destination and apply a buffer if the route contains obstacles. Notice the buffer increase for the obstacles between stage 1 and stage 2. This indicates GPT-4 recognized the second task of travelling to the final goal was more hazardous, resulting in the declaration of an increased buffer size to force A* to calculate a safer route and avoid contact with the obstacles.  }
\label{fig:fig4}
\end{figure*}

\subsection{Sequential Reasoning with Dynamic Buffer}
The third experiment shown above in Figure 4 combines sequential planning and dynamic obstacle buffering. The task involves two stages: the robot must first collect a “resource” near one goal and then navigate to a final goal which was surrounded by obstacles and more difficult to reach (see appendix A). The environment for these trials included:

\begin{itemize}
\item Two goal markers: one labeled as a resource (e.g., a goal near a long skinny obstacle) and the other as the final goal.
\item Some obstacles, including potentially a “toxic” obstacle or a narrow passage that would require careful navigation.
\item We explicitly instruct GPT-4 about the two-stage mission (as in the prompt described earlier). The expectation is that GPT-4 will direct the robot to the resource goal in stage 1, then to the final goal in stage 2.
\end{itemize}

Within this scenario, we test the dynamic buffer adjustment capability. In some trials, the direct path to the final goal passes through a congested area (e.g., between two obstacles with a tight gap). We expect GPT-4 to command a buffer increase (e.g., 20 cm) for that leg, forcing A* to find a wider path (perhaps around a different side) to maintain clearance. In other trials where the path is open, GPT-4 should use the default buffer (0 cm) to allow the shorter route. We ran 10 trials with varying obstacle configurations: some requiring the buffered path for safety, others not. We measure overall task success rate (completing both stages without collision and correctly collecting the resource first) and qualitatively note whether GPT-4’s buffer decisions align with the safety needs of the scenario.

Each experiment is logged with detailed data (timing, success/failure, path choices). For Experiment 1, we present comparative metrics between classical A* and GPT-4-guided navigation. For Experiments 2 and 3, we focus on success rates and qualitative behavior since there is no direct classical counterpart that can follow the semantic instructions.

\section{Results}

\subsection{Performance: A* vs GPT-4 Guided Planning}
Table 1 summarizes the results on the obstacle courses for both the classical A* and the GPT-4 path planning tested independently. In terms of pure navigation efficiency, the classical A* had an advantage: it found the shortest path and the robot executed it with minimal delay. On Course 1 (single obstacle), A* took on average 16.4 s to reach the goal with a 100\% success rate (no collisions in 5/5 trials). GPT-4-guided navigation on the same course took a longer total time of about 42.2 s on average, and succeeded without collision in 60\% of trials (3/5). The increased time for GPT-4 runs is attributed largely to the planning time – GPT-4’s response latency varied, averaging  15.9 s per query on Course 1. In one trial, GPT-4 took over 20 s to respond, leading to a total traversal time of 60.2 s. In terms of path quality, GPT-4 sometimes creates a path significantly longer than required or generates a path dangerously close to obstacles, which in two instances led to the robot brushing an obstacle (those were counted as failures). However, GPT-4 did not make these mistakes with any degree of predictability, making it difficult to tune prompts.

On Course 2 (two obstacles), A* again performed robustly: 16.7 s average completion, 80\% success (one trial the robot clipped an obstacle corner, resulting in a penalty). GPT-4 averaged 49.2 s and also 60\% success rate on Course 2. Notably, GPT-4’s planning times increased on some runs with more obstacles – in one trial it spent over a minute reasoning (perhaps debating path safety or generating new routes due to the collision checker deeming the route invalid), which, although resulted in a safe path, was very slow. Course 3 (three obstacles) presented a winding path challenge: A* managed 18.0 s average time with 100\% success (5/5). GPT-4, interestingly, improved to 80\% success rate on Course 3 and reduced its average planning time to 31.6 s. We suspect that by Course 3, the prompt conditioned GPT-4 to be more decisive (or the optimal path was clearer to the LLM given the metrics). Still, GPT-4’s total time was 54.0 s, roughly 3× slower than A*.

\begin{figure}[!ht]
\captionsetup{type=table}
\raggedright
\caption{COMPARISON OF CLASSICAL A* VS GPT-4 AUGMENTED PLANNING ON OBSTACLE COURSES}
\label{tab:table1}
\includegraphics[width=\linewidth]{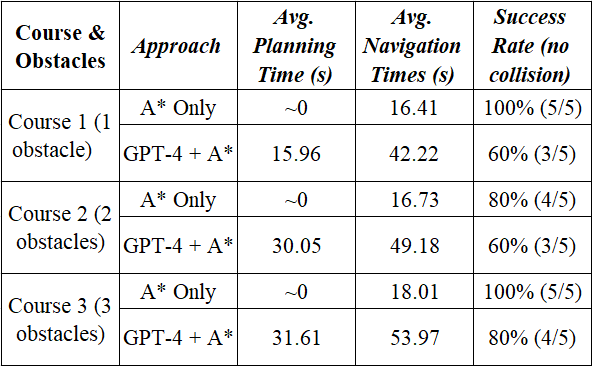}
\end{figure}

Overall, these results indicate that for straightforward obstacle avoidance tasks, our GPT-4 approach is functional but slower and significantly less reliable than a well-tuned classical planner. This lack of reliability was expected and was the reason for the implementation of a collision checker on the GPT-4 produced paths. The benefit of GPT-4, however, is not evident in pure geometry – it becomes apparent in the following semantic tests where A* alone cannot satisfy the requirements at all.

\subsection{Semantic Understanding and Navigation}
The GPT-4 integrated system excelled in following high-level instructions that were impossible for standard A* without significant alteration to tailor the path planner for each scenario as can be seen in Table 2. In the Toxic Obstacle scenario, GPT-4 correctly avoided the toxic area in all 10 trials (100\% semantic success). In practice, GPT-4 frequently chose a path that went to the goal that was clear of the toxic obstacle, even if a slightly shorter path existed on to the goal that had the toxic obstacle near it, demonstrating that it prioritized safety as instructed. The robot successfully reached the goal each time without coming close to the toxic object. In contrast, a classical A* planner (if unaware of toxicity) would simply generate a route and go to either goal with no semantic understanding guiding decisions. Thus, GPT-4 provided a capability that A* lacks – context-aware avoidance.

For the Battery (Low Battery) scenario, GPT-4 had a 90\% success rate in executing the intended behavior. In 9 out of 10 trials, GPT-4 identified that the presence of a “battery” near one goal meant the robot should go there first (to presumably “recharge”) before heading to the other goal. Interestingly, the setup was identical to the toxic obstacle scenario, with the only adjustment being that the long skinny object was described as a battery and not toxic. GPT-4 reinterpreted this obstacle as now the battery and went to that goal with relative consistency. One failure case occurred when the two goals were very close together; GPT-4 got somewhat confused and went straight to the goal without a “battery”, highlighting potential issues with system precision in congested spaces. This was easily fixed by clarifying the prompt and slightly spacing the goals further apart, after which GPT-4 behaved correctly in similar setups. A* alone would have no way to incorporate “low battery” concerns and would head to either goal without discretion, so we see this as a strong validation of GPT-4’s semantic reasoning utility.

In the Crowded Area scenario, GPT-4 also achieved 100\% success (10/10 trials) in avoiding the designated crowded region. Without explicit coding, GPT-4 inferred that “crowded” implies the robot should not go through that area. It treated the region as an unsafe zone and would opt to go to the goal which was not congested, although A* had a valid route to both goals. The robot took alternate routes around the crowd every time, at the cost of a small path length increase, which is a desirable trade-off for respecting social constraints. It’s important to highlight that in these trials the occupancy grid did not mark the crowded area in any special way, so a normal planner would once again travel to either goal without considering nearby obstacle count unless specifically coded to do so. GPT-4’s reasoning created an implicit crowded zone based on obstacle location and density surrounding goals.

Across all three semantic tests, the average semantic instruction compliance was roughly 96.7\%. The only errors were minor prompt misinterpretations that were corrected with prompt tuning. These results demonstrate GPT-4’s ability to inject high-level knowledge and constraints into the planning process effectively. 

\begin{figure}[!ht]
\captionsetup{type=table}
\raggedright
\caption{SEMANTIC TASK SUCCESS RATES WITH GPT-4 GUIDED NAVIGATION}
\label{tab:table2}
\includegraphics[width=\linewidth]{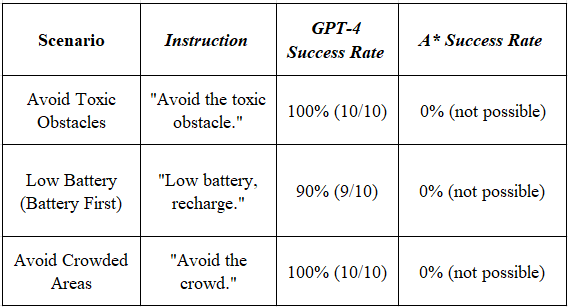}
\end{figure}

Since classical A* has no mechanism to interpret or enforce these semantic instructions, it would ignore them, giving 0\% compliance if left unmodified. GPT-4’s ability to incorporate semantic knowledge shows the real benefit of this approach.

\subsection{Sequential Tasks and Dynamic Buffer Adjustment}
 In the two-stage resource collection experiments, the GPT-4 integrated system succeeded in 100\% of trials (10/10) in terms of completing the full mission (get resource, then final goal) and respecting safety constraints. In all trials, GPT-4 correctly directed the robot to the resource first and the final goal second. This showcases multi-step reasoning – the system essentially plans an optimal subgoal order on its own, something we did not explicitly code but emerged from the prompt and GPT-4’s understanding of the task.

GPT-4 made intelligent use of the buffer parameter when needed. In trials where the final leg had a narrow corridor between obstacles or needed to walk around obstacles, GPT-4 set the buffer to 20 in its JSON output. As a result, the A* planner recomputed the path using the inflating obstacle dimensions, which yielded a path that went around the outside of both obstacles rather than between them. This eliminated the tight squeeze and significantly reduced the risk of collision. In trials where the straightforward path was already unobstructed or had ample clearance, GPT-4 kept the buffer at 0, allowing the robot to take the direct route without unnecessary detours. We thereby see GPT-4 effectively playing the role of a context-aware path optimizer: it can decide to trade off a bit of path length for much safer clearance when the situation warrants it.

To illustrate, in one congested scenario, the direct A* path we generated (with buffer 0) went between two obstacles with only $\sim$5 cm clearance on each side of the robot – a risky path that likely would result in a collision given Bittle’s width and imprecision with movement. GPT-4’s analysis noted multiple obstacles and turns, and it responded with "buffer": 20. The buffered grid’s A* solution went around the other side of the obstacle, a route that was longer but never came closer than 15 cm to any obstacle. The robot took that route and reached the goal cleanly. Without GPT-4, if we had kept buffer 0, the robot likely would have attempted the narrow path and struck an obstacle (confirming GPT-4’s decision was correct). While it is possible to hard-code a buffer in A* for obstacles, our argument is that such actions limit the flexibility of robotic behavior, especially in situations that require semantic understanding.

All 10 sequential trials had no collisions or missed objectives with the GPT-4 system. This contrasts with a purely procedural approach: had we coded a fixed routine “go to resource then goal,” it would still require a rule for when to inflate obstacles or not. GPT-4 managed to cover that logic with the same prompt reasoning, adjusting its strategy per trial conditions. While GPT-4 was not timed for reasoning during the semantic runs, it was notably fast at deciding which goal to go to and issuing the selected path. This observable decrease in time from the API call to action is likely because making a decision is much simpler to the LLM than generating a route.

\section{Discussion}
The experimental results highlight a clear trade-off and synergy between classical planning and LLM-based reasoning in robotics. We want to clarify that different prompts were tested across separate runs, not changed in the middle of a single run. This ensures consistent conditions within each individual experiment. The classical A* algorithm is fast, optimal for short paths, and guarantees collision-free motions given the map, but it has no understanding of context beyond obstacles. GPT-4, on the other hand, enriches the system with an understanding of semantics and intent, allowing it to follow complex instructions and safety rules that are not encoded in the geometry of the map. However, this comes at the cost of increased computation time and some unpredictability.

\subsection{Benefits of the Hybrid Approach}
Our approach allowed a low-cost robot to handle scenarios that would typically require additional sensors or sophisticated programming. Instead of coding separate modules for hazard avoidance, object clearance, or task sequencing, we relied on prompt engineering to convey those requirements to GPT-4. This dramatically simplified the software development: for example, to create the “avoid X” behavior, we merely describe X in the prompt and instruct GPT-4 accordingly, rather than writing and tuning a custom reactive behavior. The flexibility of this system is another upside. We can re-purpose the same robot for a new high-level task by mostly changing the prompt and a bit of perception, rather than redesigning the entire control logic.

\subsection{Limitations}
The primary drawback observed is the increased time and occasional indecision of GPT-4. In real-time robotics, a 20–60 second planning cycle is sluggish. For applications where timing is critical (e.g., fast-moving robots or time-sensitive missions), this approach would need optimization. Possible solutions include using smaller, faster language models fine-tuned for the domain or implementing a quantized model running locally instead of using an API call. Another limitation is that GPT-4’s knowledge, while vast, is generic and not tailored to our specific robot’s dynamics. For instance, GPT-4 might not inherently know Bittle’s exact turning radius or size; we had to implicitly provide that by how we describe obstacles and buffers. If not carefully prompted, an LLM could feasibly suggest an infeasible maneuver (though in our framework, A* would filter out truly infeasible paths).

There’s also the challenge of error recovery. If GPT-4 were to make an incorrect decision (say choose the wrong goal or an insufficient buffer leading to a collision), our current system would only realize the mistake after the fact (when a collision is detected or goal not achieved). Ideally, we’d have a monitor to catch and correct LLM mistakes on the fly, perhaps through a secondary “safety judge” LLM as in SAFER \cite{27} or through more robust prompt cross-checks. In practice, our success rates were high, so we didn’t face catastrophic errors, but robustness is an important consideration for broader deployment.

\subsection{Low-Cost Platform Feasibility}
Running such an AI-driven system on a \$300 robot (Bittle + Pi + camera) shows the democratization of robotics capabilities. All heavy computation (LLM inference) was effectively done via an API call – this could even be from a phone hotspot, meaning the robot itself remains cheap. This does introduce dependence on connectivity and cloud services, which could be a limitation in some uses. Alternatively, one could run an LLM on a more powerful but still affordable computer nearby (e.g., a laptop or Jetson Nano) if internet is an issue. The key point is that one doesn’t need an expensive onboard GPU or LIDARs to achieve context-aware navigation; the intelligence is in the model.

\subsection{Generality}
While our experiments were tailored to certain tasks, the framework is general. By simply editing the prompt, we could handle scenarios like “visit goals A, B, C in order” (multi-goal route planning) or “avoid wet floor” (if the camera can detect a wet floor sign and the computer vision model has been fine-tuned). The overhead camera approach could be replaced by an onboard camera with image understanding via GPT-4’s multi-modal variant or a smaller vision model feeding textual descriptions to GPT-4. The concept would remain: use classical planning for physical feasibility, use LLM for brains. This combination may become increasingly popular as LLMs improve and become more accessible.

In conclusion, the integration of GPT-4 with a classical planner allowed us to break free from rigid state machines and unlock semantic intelligence in robot navigation. The discussion above outlines both the promise and the areas for improvement. Next, we wrap up with final thoughts and future directions.

\section{Conclusion}
We presented a novel framework that transforms a low-cost robot into a semantically intelligent agent by integrating GPT-4 with the A* path planning algorithm. Through this hybrid approach, we achieved navigation behaviors that go beyond geometry: the robot can heed instructions like avoiding specific hazards or prioritizing certain goals, all without custom code for each scenario. Our experiments demonstrated that while classical A* is unbeatable in speed for simple tasks, the addition of GPT-4 greatly expands the robot’s capability to handle complex, high-level goals, with near-perfect success in semantic tasks.

The key takeaway is that advanced language models can serve as high-level “brains” for robots, even those with limited onboard resources. This reduces the need for extensive manual coding of scenario-specific behaviors. As LLMs continue to evolve, we anticipate even greater potential for enabling general-purpose, context-aware robotics on affordable platforms. We hope this work inspires further research in LLM-driven robotics, particularly toward robust safety, real-time performance, and multimodal integration.

\bibliographystyle{IEEEtran}
\bibliography{references}

\newpage

\section{Appendix A}
\label{ch:appendix_A}

\lstdefinelanguage{generalformat}{
  keywords={QUERY, RESPONSE},
  keywordstyle=\color{blue}\bfseries,
  sensitive=true,
  breaklines=true,
  keepspaces=true,
  showstringspaces=false,
  columns=fullflexible,
  basewidth = {.6em},
  breakindent = {0em},
  tabsize=1,
  aboveskip=0em,
  belowskip=0em,
  commentstyle=\color{purple}\ttfamily,
  stringstyle=\color{blue}\ttfamily
}

\begin{tcolorbox}[fonttitle=\small\fon{pbk}\bfseries,
fontupper=\scriptsize\ttfamily,
fontlower=\fon{put},
enhanced,
left=1pt, right=1pt, top=2pt, bottom=2pt,
title=Example LLM Prompt and Response]
\begin{lstlisting}[language=generalformat]
QUERY 1:
"You are an LLM controlling a BittleBot in a 2D grid.  The mission is to collect a resource. There are two  goals in your environment. The goal a little further  away with a single skinny obstacle is the resource goal and is your target. For this mission, you should only go to the goal with a single skinny obstacle near it and avoid the other goal. Based on the candidate path  metrics (including path length, obstacle count, clearances, and qualitative description of obstacles), select the candidate that travels to the resource marker and then decide on the appropriate occupancy grid buffer: if the path is direct with minimal obstacles, set the buffer to 0; if it requires turns or obstacle avoidance, set it to 20. Return your decision strictly in JSON format: 
{\"mode\":\"candidate_selection\", \"selected_candidate\":<index>, \"buffer\":<value>}. Do not include any extra fields."

"candidate_paths": [
   {"index": 0,
    "path_length": 0.5855240642829357,
    "obstacle_count": 0,
    "min_clearance": 0.13249999610707164,
    "avg_clearance": 0.13249999610707164},
    
   {"index": 1,
    "path_length": 0.9838379446400219,
    "obstacle_count": 0,
    "min_clearance": 0.0052999998442828655,
    "avg_clearance": 0.06183333151663343}
]

RESPONSE 1:
"{\"mode\":\"candidate_selection\", \"selected_candidate\":0, \"buffer\":0}"

    
QUERY 2:
"You are an LLM controlling a BittleBot in a 2D grid.  The mission is now to navigate to the final goal while maintaining safe clearance from obstacles. Among the candidate paths provided, select the one that ensures  the safest navigation. Based on the candidate path metrics (including path length, obstacle count, and clearances), determine whether the path is sufficiently direct (set buffer 0) or if it requires turns or  obstacle avoidance (set buffer 20). Return your decision strictly in JSON format: 
{\"mode\":\"candidate_selection\", \"selected_candidate\":<index>, \"buffer\":<value>}. Do not include any  extra fields.",

"candidate_paths": [
   {"index": 0,
    "path_length": 0.9434650722891916,
    "obstacle_count": 1,
    "min_clearance": 0.0052999998442828655,
    "avg_clearance": 0.06889999797567725}
]


RESPONSE 2:
"{\"mode\":\"candidate_selection\", \"selected_candidate\":0, \"buffer\":20}"
        
\end{lstlisting}
\end{tcolorbox}

\end{document}